# A Concurrent Fuzzy-Neural Network Approach for Decision Support Systems

Cong Tran, Ajith Abraham[*] and Lakhmi Jain,
School of Electrical and Information Engineering, University of South Australia, Adelaide, Australia
Email: cong.tran@postgrads.unisa.edu.au, L.Jain@unisa.edu.au
[*]Department of Computer Science, Oklahoma State University, Tulsa,
Oklahoma 74106-0700, USA. Email: ajith.abraham@ieee.org

*Abstract*- Decision-making is a process of choosing among alternative courses of action for solving complicated problems where multi-criteria objectives are involved. The past few years have witnessed a growing recognition of Soft Computing technologies that underlie the conception, design and utilization of intelligent systems. Several works have been done where engineers and scientists have applied intelligent techniques and heuristics to obtain optimal decisions from imprecise information. In this paper, we present a concurrent fuzzy-neural network approach combining unsupervised and supervised learning techniques to develop the Tactical Air Combat Decision Support System (TACDSS). Experiment results clearly demonstrate the efficiency of the proposed technique.

## I. INTRODUCTION

Several decision support systems have been developed mostly in various fields including medical diagnosis, business management, control system, command and control of defense, air traffic control and so on [4][14][15][16]. Usually previous experience or expert knowledge is often used to design decision support systems. The task becomes interesting when no prior knowledge is available. The need for an intelligent mechanism for decision support comes from the well-known limits of human knowledge processing. It has been noticed that the need for support for human decision makers is due to four kinds of limits: cognitive, economic factors, time and competitive demands. Several adaptive learning frameworks for constructing intelligent decision support systems have been proposed [3]. To develop an intelligent decision support system, we need a holistic view on the various tasks to be carried out including data management and knowledge management (reasoning techniques)[8][9]. The focus of this paper is to develop a Tactical Air Combat Decision Support System (TACDSS) with minimal prior knowledge, which could also provide optimal decision scores. This paper is an extension of our previous work wherein we had implemented evolutionary-fuzzy system [4] and different fuzzy inference methods learned using different learning techniques [5][6]. As shown in Figure 1, we propose a concurrent fuzzy – neural network [1] approach to cluster the decision regions and to automatically generate the decision scores using a neural network based on the developed decision cluster regions [11][12][13]. In Section 2, we introduce the some theoretical concepts on fuzzy c means clustering algorithm followed by neural network training using Levenberg-Marquardt algorithm [10] followed by some presentation of the complexity of the problem in tactical air combat environment (decision-making process) in Section 3. Experimentation results are provided in Section 4 and some conclusions are also provided towards the end.

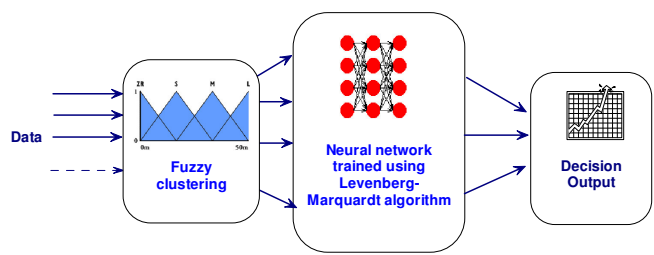

Fig. 1. Concurrent fuzzy-neural network for TACDSS

## II. COMPLEXITY OF TACDSS

The air operation division of Defence Science and Technology Organisation (DSTO), Australia and our research team has a collaborative project to develop the TACDSS for a pilot or mission commander in tactical air combat. In Figure 2 a typical scenario of air combat tactical environment is presented. The Airborne Early Warning and Control (AEW&C) is performing surveillance in a particular area of operation. It has two hornets (F/A-18s) under its control at the ground base as shown "+" in the left corner of Figure 2. An air-to-air fuel tanker (KB707) "□" is on station and the location and status are known to the AEW&C. Two of the hornets are on patrol in the area of Combat Air Patrol (CAP). Sometime later, the AEW&C on-board sensors detects 4 hostile aircrafts (Mig-29) shown as "O". When the hostile aircrafts enter the surveillance region (shown as dashed circle) the mission system software is able to identify the enemy aircraft and its distance from the Hornets in the ground base or in the CAP. The mission operator has few options to make a decision on the allocation of hornets to intercept the enemy aircraft.

- Send the Hornet directly to the spotted area and intercept,
- Call the Hornet in the area back to ground base and send another Hornet from the ground base
- Call the Hornet in the area to refuel before intercepting the enemy aircraft

The mission operator will base his decisions on a number of decision factors, such as:

- Fuel used and weapon status of hornet in the area,

- Interrupt time of Hornet in the ground base and the Hornet at the CAP to stop the hostile,
- The speed of the enemy fighter aircraft and the type of weapons it posses,
- The information of enemy aircraft with type of aircraft, weapon, number of aircraft.

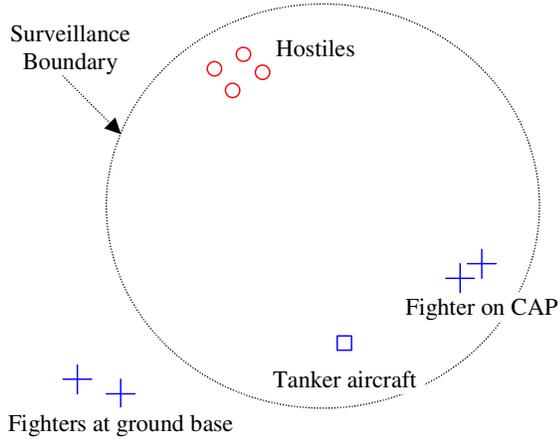

Fig. 2. A simple scenario of the air combat

From the above simple scenario, it is evident that there are several important decision factors of the tactical environment that might directly affect the air combat decision. We made use of the fuzzy neural network framework to develop the TACDSS. In the simple tactical air combat, the four decision factors that could affect the decision options of the Hornet in the CAP or the Hornet at the ground base are the following:

- *'fuel status'* - quantity of used fuel is available to perform the intercept,
- *'weapon possession status'* – quantity of weapons available in the Hornet,
- *'interrupt time'* - time required by the hornet to interrupt the hostile and
- *'danger situation'* - information of the Hornet and the hostile in the battlefield.

Each factors has difference range of unit such as the *fuel status* (0 to 1000 litres), *interrupt time* (0 to 60 minutes), *weapon status* (0 to 100 %) and *danger situation* (0 to 10 points). We used the following two expert rules for developing the fuzzy inference system.

- The decision score will have small value if the *fuel status* is low, the *interrupt time* is short, the hornet has low *weapon status*, and the *danger situation* is high.
- The decision score will have high value if the *fuel status* is full, the *interrupt time* is fast, the hornet has high *weapon status* and the *danger situation* is low.

In the air combat environment, decision-making is always based on all states of decision factors. But sometimes, a mission operator or commander could make a decision based on an important factor, such as the fuel used is too low, the enemy has more powerful weapons and the quality and quantity of enemy aircraft. Table 1 shows some typical scores (decision selection point) taking into account of the various tactical air combat decision factors.

TABLE 1
DECISION FACTORS FOR THE TACTICAL AIR COMBAT

| Fuel used | Time Intercept | Weapon Status | Danger Situation | Evaluation Plan |
|---|---|---|---|---|
| Full | Fast | Sufficient | Very Dangerous | Good |
| Half | Normal | Enough | Dangerous | Acceptable |
| Low | Slow | Insufficient | Endanger | Bad |

## III. CONCURRENT FUZZY-NEURAL NETWORK APPROACH

In a concurrent fuzzy-neural network model, the fuzzy clustering algorithm develops the required decision regions. The generated cluster information is further used to train an artificial neural network using the Levenberg-Marquardt approach to learn the decision outputs for the given input conditions. The fuzzy system continuously determines the decision regions especially when not much prior knowledge is available. Such hybrid combinations do not optimize the individual components but only aids to improve the performance of the overall system [1]. Learning takes place only in the neural network and the fuzzy system remains unchanged during this phase.

### A. Fuzzy C Means Clustering (FCM)

The problem is to perform a partition of this collection of elements into *c* fuzzy sets with respect to a given criterion, where *c* is a given number of clusters. The criterion is usually to optimise an objective function that acts as a performance index of clustering. The end result of fuzzy clustering can be expressed by a partition matrix *U* such that

$$U = [u_{ij}] \text{ with } i=1\ldots c,\ j=1\ldots n' \quad (1)$$

where $u_{ij}$ is a numerical value in [0, 1] and expresses the degree to which the element $x_j$ belongs to the $i^{th}$ cluster. There are two additional constraints on the value of $u_{ij}$. First, a total membership of the element $x_j \in X$ in all classes is equal to 1.0; that is,

$$\sum_{1}^{c} u_{ij} = 1 \quad \text{for all } j = 1, 2, \ldots, n. \quad (2)$$

Second, every constructed cluster is nonempty and different from the entire set, that is,

$$0 < \sum_{1}^{n} u_{ij} < n \quad \text{for all } i = 1, 2, \ldots, c \quad (3)$$

A general form of the *objective function* is

$$J(u_{ij}, v_k) = \sum_{i=1}^{c} \sum_{j=1}^{n} \sum_{k=1}^{c} g[w(x_i), u_{ij}] d(x_j, v_k) \quad (4)$$

where $w(x_i)$ is a weight for each $x_i$ and $d(x_j, v_k)$ is the degree of dissimilarity between the data $x_i$ and the supplemental element $v_k$, which can be considered the central vector of the $k^{th}$ cluster. The degree of dissimilarity is defined as a measure that satisfies two conditions.

(i) $d(x_j, v_d) \geq 0$,
(ii) $d(x_j, v_d) = d(v_k, x_j)$, (5)

With the above background, fuzzy clustering can be precisely formulated as an optimization problem:

Minimize $J(u_{ij}, v_k)$, $i, k = 1, 2, ..., c; j = 1, 2, ..., n$

which is subject to (2) and (3). One of the widely used clustering methods based on (5) is the fuzzy c-means (FCM) algorithm developed by Bezdek [2]. The objective function of the FCM algorithm takes the form of

$$J(u_{ij}, v_k) = \sum \sum u_{ij}^m \|x_j - v_i\|^2 \quad m > 1, \quad (6)$$

where $m$ is called the *exponential weight* which influences the degree of fuzziness of the membership (partition) matrix. To solve this minimization problem, we first differentiate the objective function in (6) with respect to $v_i$ (for fixed $u_{ij}$, $i = 1,..., c$, $j = 1,..., n$) and to $u_{ij}$ (for fixed $v_i$, $i = 1,..., c$) and apply the conditions of (2), obtaining

$$v_i = \frac{1}{\sum_{j=1}^{n}(u_{ij})^m} \sum_{j=1}^{n}(u_{ij})^m x_j \quad i = 1, 2, ..., c \quad (7)$$

$$u_{ij} = \frac{\left(\frac{1}{\|x_j - v_i\|^2}\right)^{1/m-1}}{\sum_{k=1}^{c}\left(\frac{1}{\|x_j - v_i\|^2}\right)^{1/m-1}} \quad i=1,2,...,c; j=1,2,...,n \quad (8)$$

The system described by (7) and (8) cannot be solved analytically. The FCM algorithm provides an iterative approach to approximate the minimum of the objective function starting from a given position and leads to any of its local minima [2].

*B. Neural Network*

In an artificial neural network learning occurs by the iterative updating of connection weights using a learning algorithm.

**Levenberg - Marquardt (LM) Algorithm**

When the performance function has the form of a sum of squares, then the Hessian matrix can be approximated to $H = J^T J$; and the gradient can be computed as $g = J^T e$, where $J$ is the Jacobian matrix, which contains first derivatives of the network errors with respect to the weights, and $e$ is a vector of network errors.

The Jacobian matrix can be computed through a standard backpropagation technique that is less complex than computing the Hessian matrix. The LM algorithm uses this approximation to the Hessian matrix in the following Newton-like update:

$$x_{k+1} = x_k - [J^T J + \mu I]^{-1} J^T e \quad (9)$$

When the scalar $\mu$ is zero, this is just Newton's method, using the approximate Hessian matrix. When $\mu$ is large, this becomes gradient descent with a small step size. As Newton's method is more accurate, $\mu$ is decreased after each successful step (reduction in performance function) and is increased only when a tentative step would increase the performance function. By doing this, the performance function will always be reduced at each iteration of the algorithm.

IV. MODELLING THE TACDSS

The proposed TACDSS model has 4 inputs comprising of f*uel status*, t*ime intercept*, w*eapon status* and s*ituation awareness.* The *FCM* algorithm was used to cluster the decision regions as '*good*', '*acceptable*' or '*bad*' depending on the values of the input variables. However, when the problem becomes more complex, more prior knowledge will be required to specify number of clusters (decision regions).

In complex situations, automatic clustering methods and the unsupervised fuzzy clustering [7] might be useful. The generated cluster information is made use to train an artificial neural network to learn the 3 decision regions (3 outputs) and provide the optimal decision score that will support the combat operation or commander in a battlefield situation based on the environment information.

Our master data set comprises of 1000 values representing different events (decision regions). For experimentation purposes, in order to avoid any bias on the data, from the master dataset, we randomly extracted two sets of training (Dataset A - 90% and Dataset B- 80%) and test data (10% and 20%). All the experimentations were repeated three times and the average errors are reported.

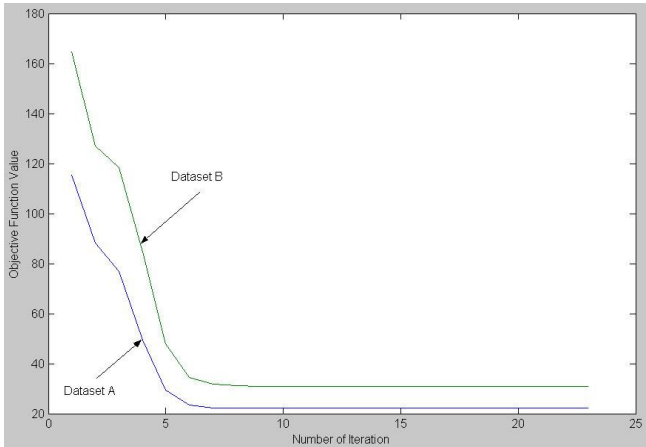

Fig. 3. Convergence of FCM algorithm for Dataset A and B

- *Experimentations Results: Fuzzy C Means Clustering*

The FCM algorithm assigns a degree of membership for the association of each data with respect to a cluster. As the iteration progresses, the cluster centers move to the 'right' position within the data set.

Figures 3 shows how the FCM algorithm converged while developing the clusters shown in Figures 4 (a) and (b) for Dataset A and B respectively. As depicted in Figure 3, for both datasets the FCM algorithm converged after nearly 15 iterations.

- *Levenberg-Marquardt Training*

We used a feed forward neural network that has 4 inputs and three target outputs. The Levenberg-Marquardt training method was used for 1500 epochs.

Figure 5 depicts the Mean Squared Error (MSE) of dataset A and B during the 1500 epochs training. The final Root Mean Squared Error (RMSE) value of dataset B (2.18E-05) was smaller value than dataset A (4.96E-05) after 1500 epochs. Figures 6 (a) and (b) show the comparison of between the actual and predicted values for training dataset A and B.

Figures 7 (a) and (b) depicts the actual and the developed TACDSS predicted decision scores for test dataset A and B. For test dataset, we obtained a RMSE of 1.0168 (dataset A) and the 0.8648 (dataset B) respectively.

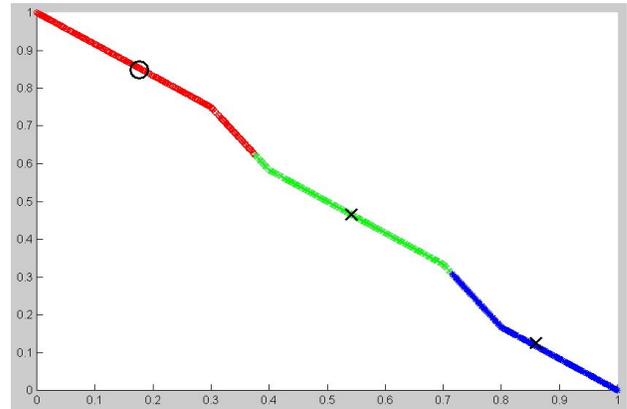
(a)

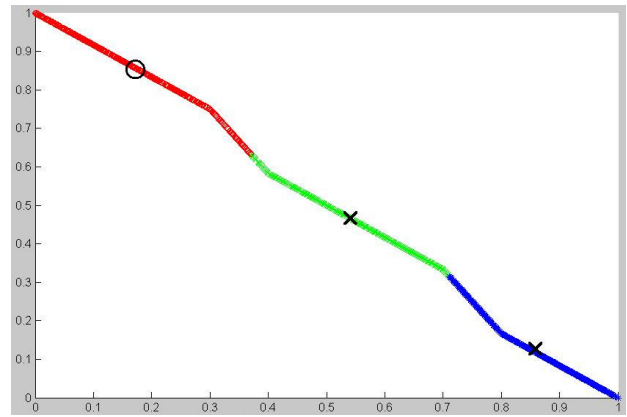
(b)

Fig. 4. Developed data clusters for dataset A and B

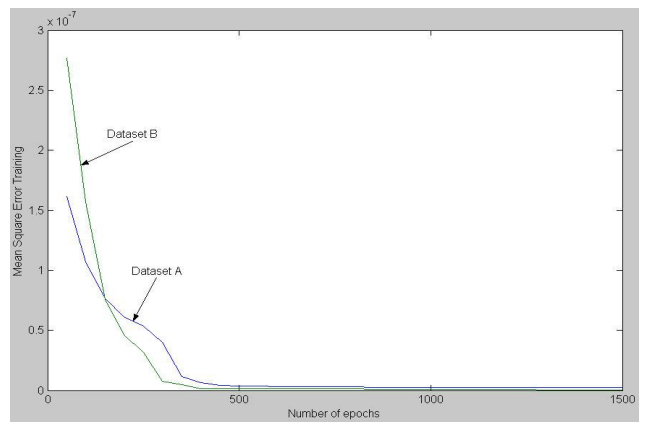

Fig. 5. Convergence of LM approach for dataset A and B

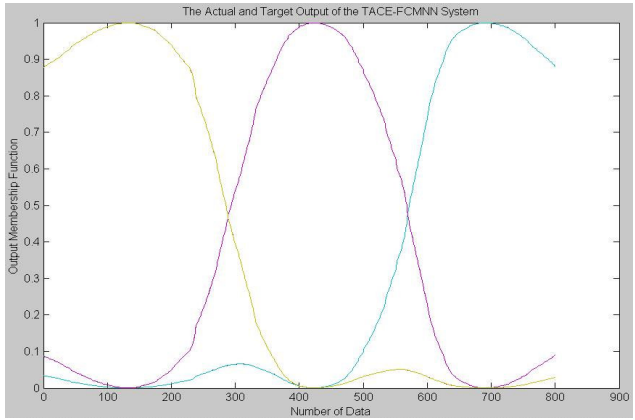

(a)

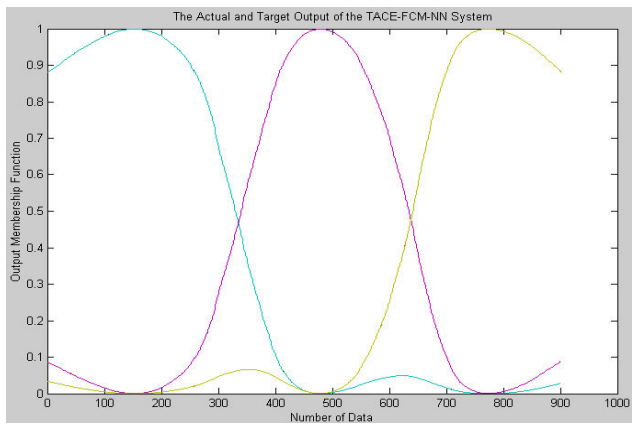

(b)

Fig. 6. Actual and predicted decision outputs for training set A and B

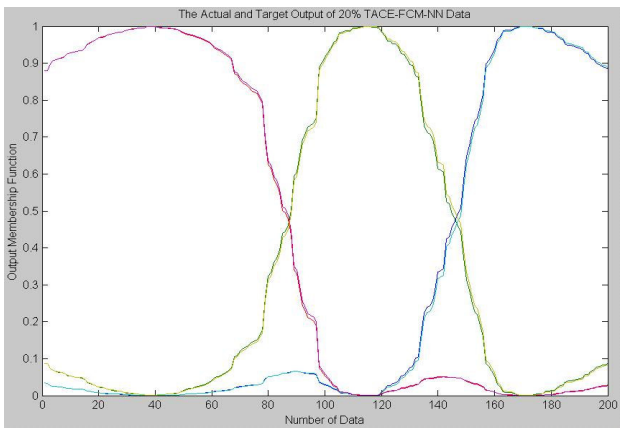

(a)

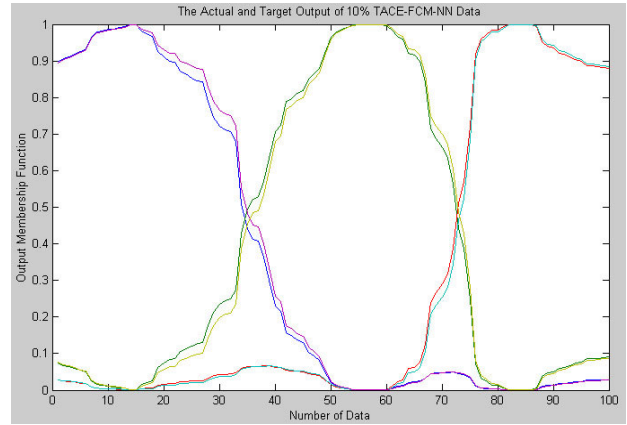

(b)

Fig. 7. Actual and predicted decision outputs for test dataset A and B

V. CONCLUSION

In this paper we proposed a method to automatically develop a decision support system using a hybrid fuzzy clustering approach and an artificial neural network working in a concurrent environment. The proposed method is very useful when not much information about the output (decision scores for given input values) is available. From the empirical results obtained, the hybrid combination of unsupervised and supervised learning seems to work well. The obtained RMSE values on test datasets are not the very best values but it is acceptable when compared with our previous methods using other connectionist paradigms, fuzzy inference systems and decision trees. In the future, we are planning to investigate the use of self-organizing maps for developing the cluster regions.